\documentclass[]{arxiv_preprint}

\usepackage{amsmath}
\usepackage{amsfonts}
\usepackage{amssymb}
\usepackage{amsthm}
\usepackage{mathtools}
\usepackage{mathrsfs}
\usepackage{nicefrac}
\usepackage{dsfont}
\usepackage{fdsymbol}
\usepackage{makecell}
\usepackage{xspace}
\usepackage{wrapfig}
\usepackage{subcaption}

\newcommand{\aidm}{LeWM\textsubscript{ActionIDM}}

\title{Is Forward Prediction Enough? Physical State Grounding for JEPA World Models}

\author[1,\ast,\dagger]{Haodong Yan}
\author[1,\ast]{Jiaguan Zhu}
\author[2]{Mingyuan Jia}
\author[2]{Ruiqing Yin}
\author[1]{Junjie He}
\author[1]{Zhide Zhong}
\author[1]{Junfeng Li}
\author[1]{Jinxuan Lu}
\author[1]{Hengtao Li}
\author[1]{Tianran Zhang}
\author[1]{Jiayi Chen}
\author[1]{Wenxuan Song}
\author[1]{Wen Chen}
\author[2]{Yuxiang Gao}
\author[1,\ddagger]{Haoang Li}

\affiliation[1]{The Hong Kong University of Science and Technology (Guangzhou), Guangzhou, China}
\affiliation[2]{COCO Matrix}
\contribution[\ast]{Equal contribution}
\contribution[\dagger]{Project Leader}
\contribution[\ddagger]{Corresponding author}

\hypersetup{
  pdftitle={Is Forward Prediction Enough? Physical State Grounding for JEPA World Models},
  pdfauthor={Haodong Yan, Jiaguan Zhu, Mingyuan Jia, Ruiqing Yin, Junjie He, Zhide Zhong, Junfeng Li, Jinxuan Lu, Hengtao Li, Tianran Zhang, Jiayi Chen, Wenxuan Song, Wen Chen, Yuxiang Gao, and Haoang Li}
}

\abstract{%
  Learning structured and control-relevant latent representations remains a key challenge for world models. Recent JEPA-based world models learn action-conditioned predictive latent dynamics from observation sequences. However, their forward-prediction objectives do not explicitly enforce reliable identifiability of robot-centric physical state from individual latents or state changes from latent pairs, which can limit downstream planning and policy performance. We propose PSG-JEPA, a physically grounded JEPA world model that shapes its latent space with two complementary grounding objectives beyond forward prediction: grounding individual latents in robot proprioceptive state, and grounding latent pairs in multi-horizon joint-angle changes. Both objectives are applied only during training, leaving the inference architecture and computational cost unchanged. To comprehensively evaluate PSG-JEPA, we conduct experiments at three levels: (1) latent identifiability via probing, (2) goal-conditioned planning on frozen latents, and (3) policy learning in simulation and on a real robot. Experiments demonstrate that our PSG-JEPA consistently outperforms state-of-the-art latent world-model baselines at all three levels.

}

\date{August 2026}
\metadata[Project Page]{\url{https://haodong-yan.github.io/psg-jepa-project-page/}}

\begin{document}

\maketitle

\begin{figure*}[t]
\centering
\includegraphics[width=\textwidth]{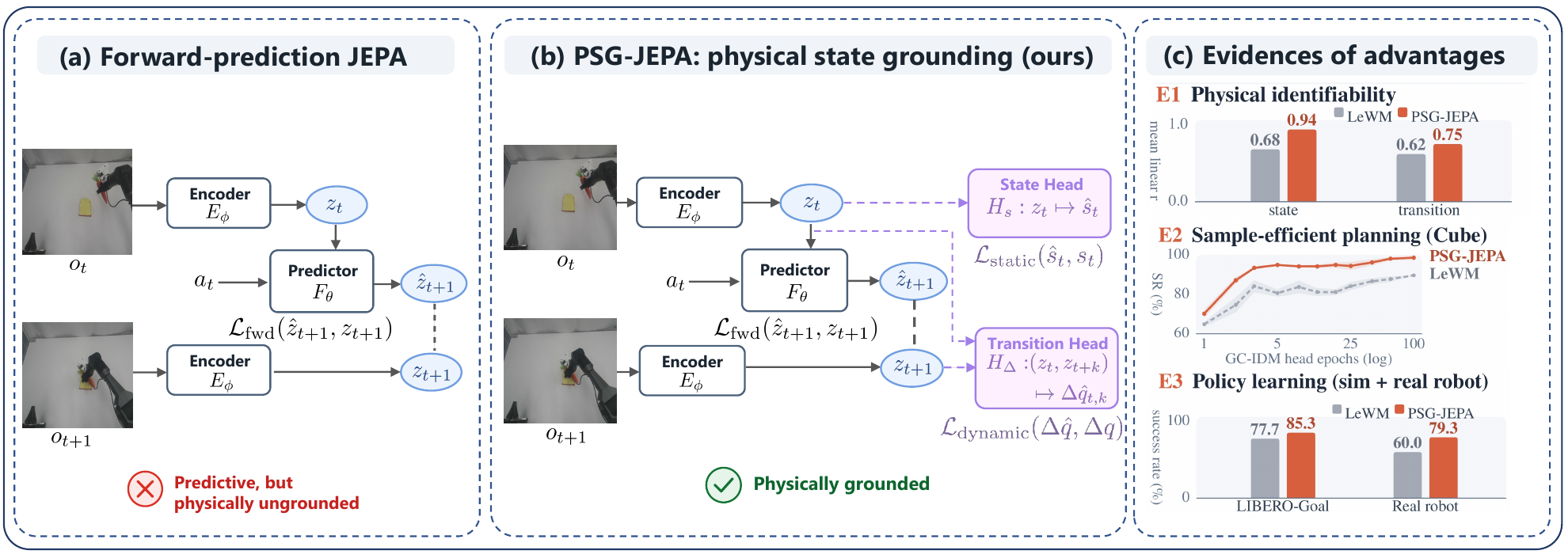}
\caption{
\textbf{Overview of the proposed physical state grounding JEPA (PSG-JEPA).}
\textbf{(a)} Forward-prediction JEPA world models (e.g., LeWM~\cite{maes2026leworldmodel}) learn a latent representation
space via action-conditioned forward prediction. However, this objective does not explicitly supervise the reliable decoding of robot-centric physical state or state change, leading to weak correlations between vision and action for downstream planners and policies.
\textbf{(b)} Our PSG-JEPA retains the action-conditioned forward pathway while
grounding individual latents in robot proprioceptive state and latent pairs in
multi-horizon joint-angle changes. These grounding objectives are implemented
through lightweight training-only state and transition heads.
\textbf{(c)} Compared to LeWM, our PSG-JEPA achieves stronger physical identifiability, higher
planning success under limited planner training, and higher policy success in simulation and on a real robot.
}
\label{fig:teaser}
\end{figure*}

\section{Introduction}

World models predict action-conditioned state transitions from observations~\cite{assran2025v,dupoux2026ai,maes2026leworldmodel,mur2026v,guo2025ctrl,ha2018world,hafner2019learning}.
Many recent world models~\cite{team2025gigaworld,liao2025genie,huang2026paiworld,guo2025ctrl} rely on video generation, with VAE latents serving as their state representations.
However, the reconstruction objective of video VAEs prioritizes pixel-level accuracy and can preserve low-level appearance details that are often unnecessary for planning or policy learning~\cite{assran2025v, nilaksh2026reconstructionsemanticsmakeslatent, nie2026lary, tong2026beyond,yan2026open}.

In contrast, latent world models forgo pixel generation and learn action-conditioned dynamics in a compact feature space~\cite{Hafner2020Dream,maes2026leworldmodel,assran2025v,zhou2024dino,lyu2026lda}.
Early latent world models learn dynamics in a fixed pretrained representation space, as in DINO-WM~\cite{zhou2024dino}, built on frozen DINOv2 features~\cite{oquab2023dinov2}, and V-JEPA 2-AC, which uses a frozen video-pretrained JEPA encoder~\cite{assran2025v}.
Such approaches benefit from strong visual priors, but physical information such as actions and robot states cannot shape the representation itself.
Recent end-to-end latent world models such as LeWM~\cite{maes2026leworldmodel} jointly optimize the encoder and the action-conditioned predictor, allowing the forward-prediction objective to shape the latent space (see Fig.~\ref{fig:teaser}(a)).
However, this objective provides no direct supervision for recovering robot-centric physical state from individual latents or state changes from latent pairs.

To quantify this limitation, we investigate forward-trained representations through frozen probe evaluations, as shown in Fig.~\ref{fig:teaser}(c): single-latent probes do not reliably recover all robot-centric physical state variables, while pairwise probes recover physical state changes only partially.
These results characterize a robot-centric identifiability gap (probe-based recoverability of physical quantities from frozen latents) under forward-only training.
Control requires relating visual observations to the robot's state and to how actions change it,
making vision--action alignment at the representation level itself a key property for
control~\cite{nie2026lary}.
When this information is not readily accessible from individual latents and latent pairs,
downstream planners and policies must learn vision--action alignment from their own task supervision.

Driven by this diagnosis, we introduce \textbf{Physical State Grounding JEPA (PSG-JEPA)}, an end-to-end JEPA world model trained with complementary state-level and transition-level grounding objectives for individual latents and latent pairs, respectively (see Fig.~\ref{fig:teaser}(b)).
First, \emph{state grounding} uses robot proprioceptive state \(s_t\) to supervise each latent \(z_t\).
Second, \emph{transition grounding} utilizes the multi-horizon joint-angle changes
\(\Delta q_{t,k}=q_{t+k}-q_t\), for all \(k\in\{1,\dots,T{-}1\}\) in the training window, to supervise
the endpoint latent pairs \((z_t,z_{t+k})\).
The two objectives constrain complementary aspects of the representation: state grounding links individual latents to robot state, while transition grounding links latent pairs to joint-angle changes across horizons.
The heads implementing these objectives are discarded after training, so PSG-JEPA introduces no inference-time overhead.
Together, the two objectives make state and state-change information readily accessible to downstream planners and policies.

In summary, our main contributions are as follows:
\begin{itemize}
    \item We characterize a robot-centric identifiability gap in forward-trained LeWM representations: not all robot-centric physical state variables are reliably recoverable from individual latents, while physical state changes are only partially recoverable from latent pairs. This gap shifts the burden of vision--action alignment onto downstream planners and policies.

    \item We propose PSG-JEPA, which addresses the above gap by grounding the latent space in robot proprioception and multi-horizon joint-angle changes, without changing the backbone architecture or adding inference-time modules.

    \item Extensive experiments show that our PSG-JEPA outperforms state-of-the-art latent world-model baselines across latent identifiability, goal-conditioned planning, and policy learning.
\end{itemize}

\section{Related Work}

\subsection{World Models}

World models learn predictive dynamics that allow agents to reason about future outcomes under candidate actions~\cite{assran2025v,dupoux2026ai,maes2026leworldmodel,mur2026v,hou2026world}.
One line of work couples world modeling with video generation, predicting future observations in RGB or in reconstruction-oriented latent spaces such as those of video VAEs~\cite{team2025gigaworld,team2026gigaworld,liao2025genie,huang2026paiworld,yang2023learning,hu2023gaia,zhao2026unidrivedreamer,song2026reconvla,zhong2025flowvla,zhong2026dualcotvla,yan2026s,yan2026open}. These models generate visually rich future observations, but reconstruction objectives preserve low-level appearance detail that is often irrelevant for planning or policy learning~\cite{assran2025v,nilaksh2026reconstructionsemanticsmakeslatent,nie2026lary,tong2026beyond}.
A second line learns compact action-conditioned latent dynamics for control, from Dreamer-style models~\cite{Hafner2020Dream,hafner2019learning,hafner2023mastering,wu2023daydreamer,zhao2025drivedreamer} to world models over semantic or predictive features. DINO-WM~\cite{zhou2024dino} plans over frozen DINOv2 features~\cite{oquab2023dinov2}, V-JEPA 2-AC adds an action-conditioned predictor on a frozen JEPA encoder~\cite{assran2025v}, and LeWM trains the encoder and predictor end to end~\cite{maes2026leworldmodel}.
In the world-model baselines above, representations are optimized for reconstruction, reward prediction, or future-feature prediction. Their objectives do not explicitly enforce reliable identifiability of either robot-centric physical state or multi-horizon state changes. PSG-JEPA targets this limitation with state- and transition-level grounding objectives for end-to-end JEPA world models.

\subsection{Joint Embedding Predictive Architecture}

Joint Embedding Predictive Architectures (JEPAs) learn non-generative representations by predicting target embeddings rather than reconstructing observations. 
I-JEPA~\cite{assran2023self} and V-JEPA~\cite{assran2025v,mur2026v} instantiate this idea for image and video representation learning, while recent action-conditioned variants adapt JEPA objectives to latent world modeling and planning~\cite{assran2025v,sobal2025stresstesting,maes2026leworldmodel}. 

Because joint-embedding objectives admit collapsed solutions, JEPA training relies on explicit anti-collapse regularization, e.g., the principled isotropic-Gaussian regularizer (SIGReg) of LeJEPA~\cite{balestriero2025lejepa}.
Some JEPA-based world models~\cite{sobal2025stresstesting} instead attach inverse-dynamics or action-decoder objectives to latent transitions, and these objectives serve a similar regularizing role.
They encourage action-discriminative representations by predicting actions from latent pairs, \((z_t,z_{t+k}) \mapsto a_{t:t+k-1}\).
However, inverse-dynamics supervision is ambiguous as a physical-transition signal: even from the same initial state, multiple action sequences can produce the same endpoint state change.
Our PSG-JEPA instead grounds latent pairs in the net joint-angle change \(\Delta q_{t,k}=q_{t+k}-q_t\), which is uniquely determined by the endpoint states and fixed-dimensional at every horizon.
This gives each latent pair a single, well-defined physical transition target, making the correspondence between latent transitions and state change more readily learnable across horizons.

\subsection{State-Aware Robot Representation Learning}

A complementary line of work uses privileged robot state to shape visual representations for downstream
robot tasks. MCR~\cite{jiang2025mcr} pre-trains a visual encoder on large robot datasets by aligning each
image feature with a temporal window of proprioceptive states and actions and by predicting actions
through a behavior-cloning head. RS-CL~\cite{kim2026rscl} regularizes a vision--language--action model at fine-tuning time so
that the latent similarity between two samples reflects their proprioceptive-state proximity.
RoboPEPP~\cite{goswami2025robopepp} pre-trains a single-image embedding-predictive encoder by masking
robot-joint regions, for robot pose and joint-angle estimation. We share their premise that the
proprioceptive state is freely available in robot data. However, none of these objectives regresses a physical transition over a temporally ordered,
same-trajectory latent pair, which is precisely what PSG-JEPA grounds by mapping each pair
\((z_{t},z_{t+k})\) to the net joint-angle change \(\Delta q_{t,k}\) at multiple horizons.
PSG-JEPA also differs in scope and evidence: it characterizes and
addresses a robot-centric identifiability limitation in end-to-end JEPA world
models, and evaluates the remedy at three levels (identifiability, planning, and
policy learning) rather than through downstream performance alone.

\section{Method}
\label{sec:method}

\begin{wrapfigure}{r}{0.48\textwidth}
\centering
\captionsetup{font=footnotesize}
\includegraphics[width=\linewidth]{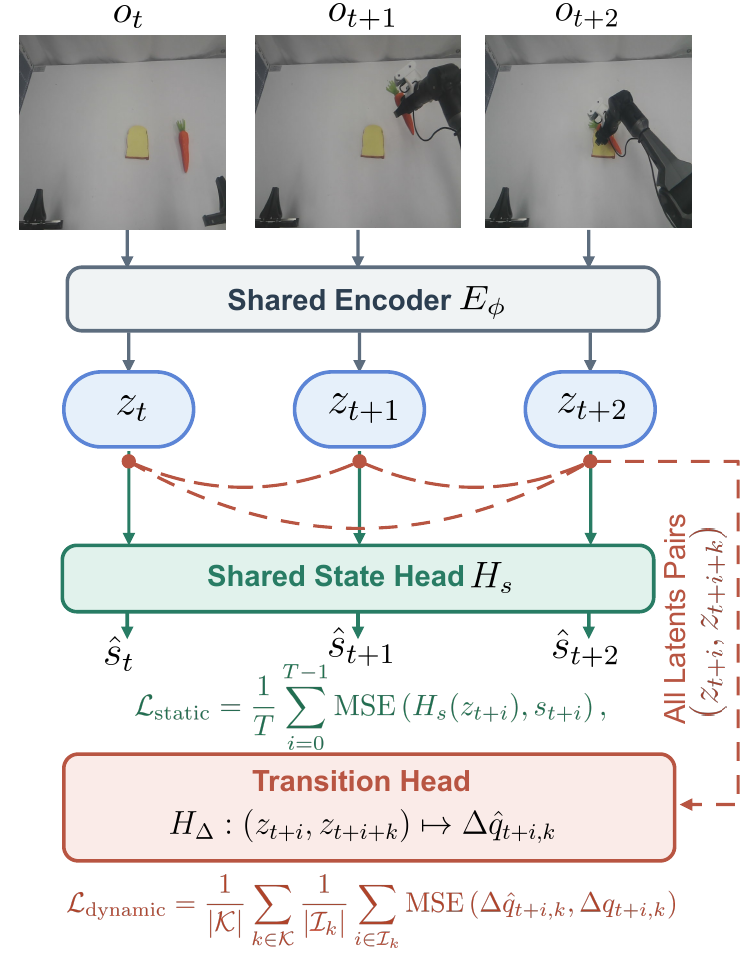}
\caption{
\textbf{Illustration of PSG-JEPA grounding objectives.}
For each image observation \(o_{t+i}\), the shared encoder produces
\(z_{t+i}=E_\phi(o_{t+i})\). \(s_{t+i}\) is the aligned logged robot state, and
\(q_{t+i}\) is its joint-angle component. The state head predicts
\(\hat{s}_{t+i}=H_s(z_{t+i})\) from each latent, while the transition head predicts
\(\Delta\hat{q}_{t+i,k}=H_\Delta(z_{t+i},z_{t+i+k})\) from every valid endpoint pair,
supervised by \(\Delta q_{t+i,k}=q_{t+i+k}-q_{t+i}\). Here, \(i\) indexes the
starting frame and \(k\) the horizon. The figure shows a three-frame example. A
\(T\)-frame window uses all \(k=1,\dots,T{-}1\).
}
\label{fig:method}
\end{wrapfigure}

We introduce \textbf{PSG-JEPA}, a physical state grounding framework for end-to-end JEPA world models.
As illustrated in Fig.~\ref{fig:teaser}(b), PSG-JEPA preserves the forward JEPA prediction pathway while introducing two complementary physical grounding objectives: \emph{static state grounding} for individual latents and \emph{dynamic transition grounding} for latent pairs.
The grounding heads are used only during training and discarded during inference. 
We first describe the retained forward JEPA world model (Sec.~\ref{sec:method-forward}), then
introduce static state grounding (Sec.~\ref{sec:method-static}) and dynamic transition grounding
(Sec.~\ref{sec:method-dynamic}), and finally present the overall training objective
(Sec.~\ref{sec:method-objective}).

\subsection{Preliminaries: Forward JEPA World Model}
\label{sec:method-forward}

PSG-JEPA retains the JEPA encoder and causal action-conditioned predictor. For a training window with \(T=C+1\)
observations and \(C\) actions, the encoder maps each observation to
\(z_{t+i}=E_\phi(o_{t+i})\) for \(i=0,\dots,T{-}1\), and the action-conditioned predictor \(F_\theta\) uses the first \(C\)
latent--action pairs to produce teacher-forced one-step predictions:
\begin{equation}
\hat z_{t+1:t+C}
=
F_\theta\!\left(z_{t:t+C-1},a_{t:t+C-1}\right),
\end{equation}
where each prediction \(\hat z_{t+i+1}\) attends only to the prefix through time \(t+i\). The forward objective is:
\begin{equation}
\mathcal{L}_{\mathrm{fwd}}
=
\frac{1}{C}\sum_{i=0}^{C-1}
\operatorname{MSE}\!\left(\hat z_{t+i+1},z_{t+i+1}\right).
\end{equation}
Following LeWM~\cite{maes2026leworldmodel}, we retain its SIGReg~\cite{balestriero2025lejepa} anti-collapse regularizer without
modification. The resulting forward JEPA objective is:
\begin{equation}
\mathcal{L}_{\mathrm{JEPA}}
=
\mathcal{L}_{\mathrm{fwd}}
+
\lambda_{\mathrm{reg}}\mathcal{L}_{\mathrm{SIGReg}},
\end{equation}
where \(\lambda_{\mathrm{reg}}=0.09\) is the regularization weight.
This objective makes future features predictable conditioned on actions, but does not explicitly ground individual latents in robot state or latent pairs in physical state change.
PSG-JEPA adds the two grounding objectives to impose these complementary constraints.

\subsection{Static State Grounding}
\label{sec:method-static}

As shown in Fig.~\ref{fig:method}, for a \(T\)-frame training window, a lightweight state head \(H_s\) grounds every latent through:
\begin{equation}
\mathcal{L}_{\mathrm{static}}
=
\frac{1}{T}\sum_{i=0}^{T-1}\operatorname{MSE}\!\left(H_s(z_{t+i}), s_{t+i}\right),
\end{equation}
where \(s_t \in \mathbb{R}^{d_s}\) denotes the robot proprioceptive state, including joint angles, gripper state, and end-effector pose.
This supervision makes robot proprioceptive state more readily identifiable from individual latents. Because \(H_s\) reads from the shared planning latent \(z_{t+i}\) rather than a separate branch, minimizing \(\mathcal{L}_{\mathrm{static}}\) encourages the encoder itself to expose physical state in the representation used downstream. Because physical state is a property of a single instant, static grounding operates on individual latents, complementing the pairwise transition grounding in Sec.~\ref{sec:method-dynamic}.

\subsection{Dynamic Transition Grounding}
\label{sec:method-dynamic}

As illustrated in Fig.~\ref{fig:method}, dynamic grounding complements static grounding
with a shared transition head that directly predicts joint-angle changes across horizons.
Although the joint angles are part of \(s_t\), static grounding supervises each endpoint independently and
does not directly optimize a pairwise readout of their change. A natural alternative is to supervise each pair with the intervening action sequence, as in inverse-dynamics objectives. However, this target is ill-posed as a physical-transition signal: its dimension grows with the horizon, and different action sequences can drive the robot between the same two configurations, so the same latent pair may map to many valid targets. In contrast, the overall joint-angle change between the two endpoints is uniquely determined by the endpoint states and keeps a fixed dimension across all horizons.

This motivates grounding latent pairs in the joint-angle change rather than the action sequence.
For each horizon \(k \in \mathcal{K}=\{1,\dots,T{-}1\}\), let
\(\mathcal{I}_k=\{0,\dots,T{-}k{-}1\}\) denote the valid starting offsets. For each
\(i\in\mathcal{I}_k\), we define the \(d_q\)-dimensional joint-angle change:
\begin{equation}
\Delta q_{t+i,k} = q_{t+i+k} - q_{t+i}.
\end{equation}
Here, \(q_t\in\mathbb{R}^{d_q}\) is the vector of joint angles in \(s_t\).
A transition head predicts
\(\Delta\hat{q}_{t+i,k}=H_\Delta(z_{t+i},z_{t+i+k})\), with loss:
\begin{equation}
 \mathcal{L}_{\mathrm{dynamic}}
  = \frac{1}{|\mathcal{K}|}\sum_{k\in\mathcal{K}}
    \frac{1}{|\mathcal{I}_k|}\sum_{i\in\mathcal{I}_k}
    \operatorname{MSE}\!\left(\Delta\hat{q}_{t+i,k},\, \Delta q_{t+i,k}\right).
\end{equation}
All valid pairs contribute, and grounding every horizon rather than only adjacent pairs makes both short- and long-range physical change readable from latent endpoints. Because short horizons yield more pairs than long ones, we weight each horizon equally rather than each pair, so short-range transitions do not dominate the objective.

\subsection{Overall Training Objective}
\label{sec:method-objective}

The PSG-JEPA objective combines the retained forward JEPA objective with static and dynamic physical grounding:
\begin{equation}
\mathcal{L}_{\mathrm{PSG}}
=
\mathcal{L}_{\mathrm{JEPA}}
+
\lambda_{\mathrm{g}}\!\left(
\mathcal{L}_{\mathrm{static}}
+
\mathcal{L}_{\mathrm{dynamic}}
\right),
\end{equation}
where \(\lambda_{\mathrm{g}}=0.1\) is the grounding weight, shared by both grounding terms
and kept fixed across all benchmarks. After training, \(H_s\) and \(H_\Delta\) are discarded,
so downstream evaluations use the learned encoder without any test-time grounding heads.

\section{Experiments}

We evaluate PSG-JEPA on simulated and real-world robotic tasks, organizing the evaluation around four questions.
\textbf{(Q1)} Does physical grounding make forward-trained JEPA latents more physically identifiable?
\textbf{(Q2)} Do physically grounded representations improve goal-conditioned planning on frozen latents?
\textbf{(Q3)} Do physically grounded representations improve policy learning in simulation and on real robots?
\textbf{(Q4)} How does each grounding component contribute?

\subsection{Experimental Setup}
\label{sec:setups}

To ensure a controlled comparison, our PSG-JEPA retains exactly the JEPA encoder--predictor backbone of
LeWM~\cite{maes2026leworldmodel}. The model takes image observations and learns an encoder \(E_\phi\)
and an action-conditioned predictor \(F_\theta\). We use \(C=3\) context frames and
single-step prediction, giving training windows of \(T=4\) frames and grounding horizons
\(\mathcal{K}=\{1,2,3\}\).
PSG-JEPA and LeWM therefore differ \emph{only} in the training-time grounding objectives applied to the latent,
and the proprioceptive quantities serve only as grounding targets or probe labels, never as inputs to
the encoder or predictor.

\subsection{Methods for Comparison}
\label{sec:methods}

We compare PSG-JEPA against three baselines. \textbf{LeWM}~\cite{maes2026leworldmodel} is the
same-backbone forward-prediction baseline (SIGReg regularization, no grounding heads).
\textbf{\aidm} is a variant of LeWM that we construct by adding a one-step
action-prediction (inverse-dynamics) objective on adjacent latent pairs,
\((z_t,z_{t+1})\mapsto a_t\). It tests whether action-level supervision, the natural
alternative to our physical state grounding (Sec.~\ref{sec:method-dynamic}), is
sufficient for the same benefits. \textbf{DINOv2}~\cite{oquab2023dinov2} is a pretrained
visual reference evaluated under the same downstream protocol: frozen features for
Q1--Q2, and a DINOv2-base encoder fine-tuned with the same policy head for Q3.

\subsection{Physical Identifiability Probes (Q1)}
\label{sec:probe}

We first assess whether physical grounding improves latent identifiability. Following
LeWM~\cite{maes2026leworldmodel}, we freeze the encoder and fit linear ridge and shallow
MLP probes using an episode-level train/test split. They measure linear and nonlinear recoverability,
respectively. Single-latent probes predict robot proprioception from \(z_t\). Pairwise probes
predict joint velocity, gripper velocity, or action from \((z_t,z_{t+1})\). We report Pearson \(r\)
on held-out test episodes.

\subsubsection{Robot Proprioception Identifiability.}
\begin{wraptable}{r}{0.48\textwidth}
\centering
\captionsetup{font=footnotesize}
\caption{
\textbf{Single-latent state probes} on OGBench-Cube~\cite{park2025ogbench} proprioceptive quantities.
Cells report linear\,/\,MLP Pearson \(r\) from frozen planning latents. Our PSG-JEPA shows stronger identifiability of robot proprioception.
}
\label{tab:state_probe_main}
\scriptsize
\setlength{\tabcolsep}{2.4pt}
\resizebox{\linewidth}{!}{%
\begin{tabular}{lcccc}
\toprule
Method & JointPos & EEPos & Gripper & EE-yaw \\
\midrule
LeWM & 0.71\,/\,0.69 & 0.99\,/\,\textbf{0.99} & 0.93\,/\,0.96 & 0.08\,/\,0.08 \\
DINOv2        & 0.73\,/\,0.72 & 0.99\,/\,0.94 & 0.84\,/\,0.84 & 0.51\,/\,0.50 \\
\aidm{} & 0.75\,/\,0.72 & \textbf{1.00}\,/\,\textbf{0.99} & 0.96\,/\,\textbf{0.98} & 0.11\,/\,0.10 \\
\midrule
\textbf{PSG-JEPA (ours)} & \textbf{0.83}\,/\,\textbf{0.81} & \textbf{1.00}\,/\,\textbf{0.99} & \textbf{0.97}\,/\,\textbf{0.98}
                  & \textbf{0.94}\,/\,\textbf{0.98} \\
\bottomrule
\end{tabular}%
}
\end{wraptable}
As shown in Table~\ref{tab:state_probe_main}, forward-only training does not make robot
orientation readily identifiable from a single latent. LeWM and \aidm{} leave EE-yaw
nearly unreadable (\(r\le0.11\)). Even with large-scale visual pretraining, DINOv2 recovers
EE-yaw only partially (MLP \(r=0.50\)), whereas our PSG-JEPA reaches \(r=0.98\). This gap shows
that explicit state grounding makes orientation substantially more accessible in the latent. State
grounding also improves the linear recoverability of joint angles relative to LeWM
(\(r: 0.71\to0.83\)).

\subsubsection{Transition Identifiability.}
\begin{wraptable}{r}{0.48\textwidth}
\centering
\captionsetup{font=footnotesize}
\caption{
\textbf{Transition probes} on OGBench-Cube~\cite{park2025ogbench} from adjacent latent pairs
\((z_t,z_{t+1})\). Cells report linear\,/\,MLP Pearson \(r\). Our PSG-JEPA shows stronger transition identifiability.
}
\label{tab:transition_probe_main}
\scriptsize
\setlength{\tabcolsep}{3pt}
\resizebox{\linewidth}{!}{%
\begin{tabular}{lccc}
\toprule
Method & JointVel & GripVel & Action \\
\midrule
LeWM & 0.68\,/\,0.66 & 0.44\,/\,0.47 & 0.74\,/\,0.76 \\
DINOv2        & 0.51\,/\,0.39 & 0.20\,/\,0.12 & 0.54\,/\,0.45 \\
\aidm{} & 0.73\,/\,0.69 & 0.67\,/\,0.73 & \textbf{0.80}\,/\,0.84 \\
\midrule
\textbf{PSG-JEPA (ours)} & \textbf{0.75}\,/\,\textbf{0.75} & \textbf{0.69}\,/\,\textbf{0.76}
                  & \textbf{0.80}\,/\,\textbf{0.86} \\
\bottomrule
\end{tabular}%
}
\end{wraptable}
We next use transition probes to assess whether physical state change can be recovered from latent
pairs, as shown in Table~\ref{tab:transition_probe_main}. Forward-only training leaves this information only
partially accessible. We evaluate this through velocities and actions, which describe
changes between frames. LeWM makes gripper velocity only weakly accessible (linear
\(r=0.44\)). Our PSG-JEPA matches or exceeds all baselines on each transition quantity.
In particular, our PSG-JEPA matches \aidm{} in action recoverability despite not directly
using action prediction as an auxiliary objective. Even with large-scale visual
pretraining, DINOv2 recovers little gripper velocity
(0.20\,/\,0.12), whereas our PSG-JEPA reaches 0.69\,/\,0.76 without using gripper
velocity as an explicit auxiliary target. Together, these results support the
state/transition decomposition: physical state should be accessible from
individual latents, and physical change from latent pairs.

\subsection{Goal-Conditioned Planning on Frozen Latents (Q2)}
\label{sec:planning}

\begin{wraptable}[27]{R}{0.48\textwidth}
\centering
\captionsetup{font=footnotesize}
\caption{
\textbf{Planning success rate (\%)} on OGBench-Cube and OGBench-Scene~\cite{park2025ogbench}. Each method
uses the same GC-IDM~\cite{nguyen2026latent} planning head trained on frozen latents with full or \(25\%\)
demonstration data (mean \(\pm\) std over 3 planner seeds).
}
\label{tab:planning}
\scriptsize
\setlength{\tabcolsep}{1.5pt}
\renewcommand{\arraystretch}{0.88}

{\sffamily\bfseries OGBench-Cube, full data\par}
\resizebox{\linewidth}{!}{%
\begin{tabular}{lcccc}
\toprule
Epochs & 5 & 10 & 25 & 100 \\
\midrule
LeWM & 80.7\(\pm\)1.9 & 83.3\(\pm\)0.8 & 84.2\(\pm\)2.1 & 89.7\(\pm\)0.2 \\
DINOv2 & 76.5\(\pm\)1.6 & 80.5\(\pm\)1.1 & 85.3\(\pm\)1.3 & 92.0\(\pm\)0.8 \\
\aidm{} & 82.8\(\pm\)1.2 & 87.3\(\pm\)1.3 & 94.0\(\pm\)0.7 & 96.0\(\pm\)0.4 \\
\textbf{PSG-JEPA} & \textbf{95.0\(\pm\)0.7} & \textbf{92.7\(\pm\)1.9} & \textbf{94.5\(\pm\)2.1} & \textbf{98.7\(\pm\)1.2} \\
\bottomrule
\end{tabular}%
}

\smallskip
{\sffamily\bfseries OGBench-Cube, \(25\%\) data\par}
\resizebox{\linewidth}{!}{%
\begin{tabular}{lcccc}
\toprule
Epochs & 5 & 10 & 25 & 100 \\
\midrule
LeWM & 67.0\(\pm\)0.7 & 80.8\(\pm\)2.7 & 81.7\(\pm\)2.4 & 83.5\(\pm\)3.9 \\
DINOv2 & 60.8\(\pm\)5.1 & 70.0\(\pm\)0.8 & 79.8\(\pm\)1.0 & 85.8\(\pm\)1.7 \\
\aidm{} & 63.2\(\pm\)1.8 & 77.5\(\pm\)1.5 & 82.5\(\pm\)1.9 & 91.5\(\pm\)1.1 \\
\textbf{PSG-JEPA} & \textbf{76.7\(\pm\)2.2} & \textbf{88.2\(\pm\)1.3} & \textbf{93.0\(\pm\)1.8} & \textbf{93.2\(\pm\)0.2} \\
\bottomrule
\end{tabular}%
}

\smallskip
{\sffamily\bfseries OGBench-Scene, full data\par}
\resizebox{\linewidth}{!}{%
\begin{tabular}{lcccc}
\toprule
Epochs & 5 & 10 & 25 & 100 \\
\midrule
LeWM & 76.2\(\pm\)1.2 & 83.3\(\pm\)1.3 & 87.2\(\pm\)1.0 & 91.0\(\pm\)1.1 \\
DINOv2 & 72.2\(\pm\)1.9 & 73.2\(\pm\)1.0 & 83.2\(\pm\)1.2 & 90.2\(\pm\)1.4 \\
\aidm{} & 80.3\(\pm\)0.6 & 85.8\(\pm\)0.6 & 90.5\(\pm\)0.4 & 94.7\(\pm\)0.6 \\
\textbf{PSG-JEPA} & \textbf{83.5\(\pm\)2.4} & \textbf{88.2\(\pm\)2.4} & \textbf{93.2\(\pm\)1.3} & \textbf{96.2\(\pm\)0.8} \\
\bottomrule
\end{tabular}%
}

\smallskip
{\sffamily\bfseries OGBench-Scene, \(25\%\) data\par}
\resizebox{\linewidth}{!}{%
\begin{tabular}{lcccc}
\toprule
Epochs & 5 & 10 & 25 & 100 \\
\midrule
LeWM & 62.2\(\pm\)1.2 & 70.0\(\pm\)1.8 & 76.7\(\pm\)0.9 & 83.0\(\pm\)1.1 \\
DINOv2 & 69.2\(\pm\)0.2 & 68.0\(\pm\)2.8 & 75.2\(\pm\)1.7 & 80.0\(\pm\)1.1 \\
\aidm{} & 65.5\(\pm\)1.1 & 74.5\(\pm\)1.4 & 80.8\(\pm\)3.0 & 87.3\(\pm\)1.6 \\
\textbf{PSG-JEPA} & \textbf{69.5\(\pm\)1.1} & \textbf{76.2\(\pm\)1.9} & \textbf{83.3\(\pm\)1.9} & \textbf{89.7\(\pm\)0.9} \\
\bottomrule
\end{tabular}%
}
\end{wraptable}

\begin{figure*}[!t]
\centering
\begin{subfigure}{0.48\textwidth}
\centering
\includegraphics[width=0.99\linewidth]{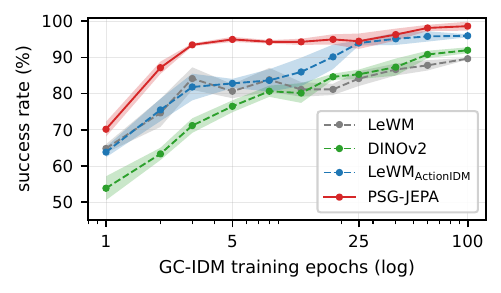}
\caption{Success rate under different GC-IDM~\cite{nguyen2026latent} training epochs, with full demonstration data.}
\label{fig:planning_epochs}
\end{subfigure}\hfill
\begin{subfigure}{0.48\textwidth}
\centering
\includegraphics[width=0.99\linewidth]{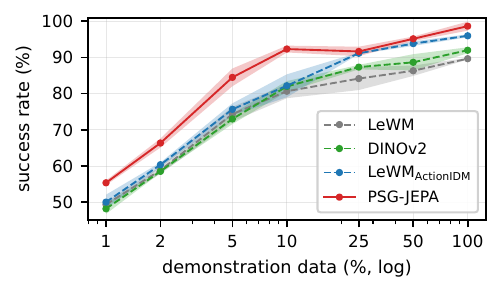}
\caption{Success rate under different demonstration-data fractions, with the planner trained for \(100\) epochs.}
\label{fig:planning_data}
\end{subfigure}
\caption{\textbf{Efficient goal-conditioned planning under limited optimization and demonstration data.} Markers indicate evaluated
points. Bands are \(\pm1\) std over \(3\) planner seeds. Our PSG-JEPA leads along both axes, with the largest margins at small planner budgets and small
data.}
\label{fig:planning_external}
\end{figure*}

We next evaluate whether physically grounded representations benefit goal-conditioned planning. For
each representation, we discard the training-time grounding heads, freeze the encoder, and train the
same GC-IDM planner~\cite{nguyen2026latent}, an amortized goal-conditioned inverse-dynamics model
implemented as a three-layer MLP with 512 hidden units and AdaLN-Zero conditioning. We evaluate
closed-loop goal reaching on OGBench~\cite{park2025ogbench}. All methods are tested on the same 200 goal instances
using the OGBench goal-reaching criterion. To comprehensively characterize planning efficiency across optimization budgets
  and data availability, we vary the GC-IDM training budget and the amount of
  available demonstration data. Table~\ref{tab:planning} reports the
  training-budget sweep at full and \(25\%\) data, while
  Fig.~\ref{fig:planning_external} further traces performance across data fractions.

\paragraph{Efficient planning under limited optimization and demonstration data.}
Under limited optimization, our PSG-JEPA reaches \(95.0\%\) success after only \(5\)
GC-IDM epochs (full data), compared with \(80.7\%\) for LeWM. Even after \(100\)
epochs, LeWM reaches only \(89.7\%\). Under limited demonstrations, with
the planner trained for \(100\) epochs, PSG-JEPA reaches \(84.5\%\) success from
only \(5\%\) of the training data, a level LeWM requires roughly five times
more data to match (see Fig.~\ref{fig:planning_external}(b)). At low GC-IDM
training budgets, both DINOv2 and \aidm{} also trail PSG-JEPA
(see Table~\ref{tab:planning}), showing that neither large-scale visual pretraining
nor action-level supervision alone yields equally efficient planning. Thus,
physical grounding makes the latent more readily usable under both limited
planner optimization and limited demonstrations.

\paragraph{Cross-environment validation on OGBench-Scene.}
To test whether these gains are specific to OGBench-Cube, we retrain the three
learned world-model encoders (DINOv2 remains frozen) and repeat the identical
protocol on OGBench-Scene~\cite{park2025ogbench}, a compound-goal manipulation task in which a cube, two
buttons, a drawer, and a window must all reach their goal configurations.
PSG-JEPA is best in every cell of the OGBench-Scene block of Table~\ref{tab:planning}.
Similar to the results on OGBench-Cube, \aidm{} is the closest baseline at
large planner budgets, while frozen DINOv2 is the hardest
representation to exploit at low planner budgets with full data. Along the data
axis, the planner trained on PSG-JEPA latents with half of the demonstrations
(\(94.7\%\)) matches or exceeds every baseline planner trained on the full dataset.

\paragraph{Open-loop latent prediction.}
\mbox{}\par
Because GC-IDM~\cite{nguyen2026latent} evaluates frozen representations without using the predictor \(F_\theta\), we
separately examine the retained predictor on OGBench-Cube and
OGBench-Scene. Starting from three observed context frames, each model
recursively predicts future latents from logged actions, without teacher forcing,
on the same \(512\) sampled trajectory segments per environment. At the shortest rollout
(\(5\) model steps, \(25\) environment steps), PSG-JEPA already attains lower latent
MSE than the same-backbone LeWM baseline, at \(0.0046\) vs.\ \(0.0093\) on Cube and
\(0.0208\) vs.\ \(0.0269\) on Scene. The advantage grows at the longest rollout
(\(30\) model steps, \(150\) environment steps), where PSG-JEPA lowers the MSE from
\(0.1488\) to \(0.0485\) on Cube (a \(67\%\) reduction) and from \(0.1608\) to
\(0.0982\) on Scene (a \(39\%\) reduction). PSG-JEPA has lower MSE at every reported horizon, so
grounding improves rather than degrades recursive predictive fidelity.

\begin{figure*}[t]
\centering
\includegraphics[width=0.99\textwidth]{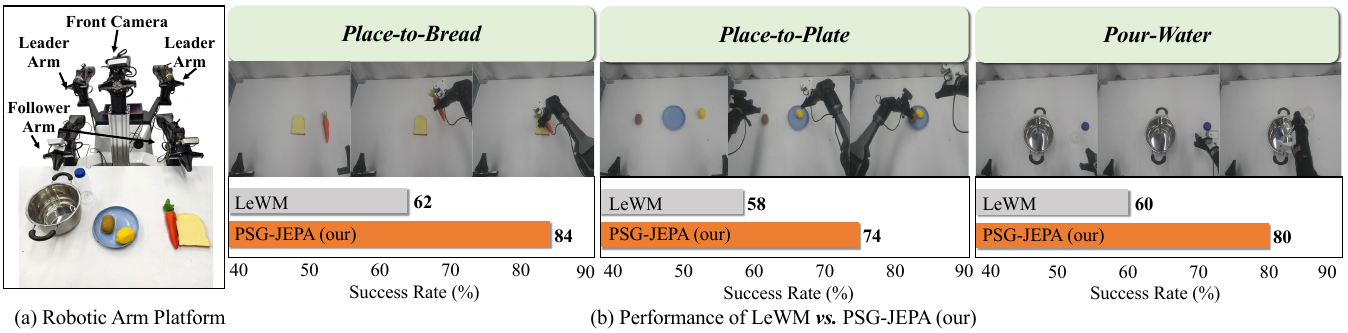}
\caption{
\textbf{Real-robot evaluation.} \textbf{(a)} Dual-arm Cobot teleoperation
platform (Mobile ALOHA design). \textbf{(b)} Per-task success rate of our PSG-JEPA
versus the LeWM baseline on Place-to-Bread, Place-to-Plate, and Pour-Water.
Our PSG-JEPA improves over LeWM on all three tasks.
}
\label{fig:real_robot}
\end{figure*}

\subsection{Policy Learning (Q3)}
\label{sec:policy}

Finally, we test whether the benefits of physical grounding extend beyond goal-conditioned planning to
policy learning, in simulation and on a real robot.

\subsubsection{Simulated Benchmarks.}

\WFclear
\begin{wraptable}[15]{r}{0.48\textwidth}
\centering
\captionsetup{font=footnotesize}
{\small
\setlength{\tabcolsep}{1mm}
\begin{tabular}{lc}
\toprule
Method & Success (\%) \\
\midrule
LeWM~\cite{maes2026leworldmodel} & 77.7\(\pm\)0.5 \\
\aidm{} & 82.6\(\pm\)2.2 \\
DINOv2~\cite{oquab2023dinov2} & 80.1\(\pm\)5.3 \\
\midrule
\textbf{PSG-JEPA (ours)} & \textbf{85.3\(\pm\)3.9} \\
\bottomrule
\end{tabular}
}
\caption{
\textbf{Policy learning on LIBERO-Goal} (mean \(\pm\) std over 3 seeds,
10 tasks \(\times\) 50 rollouts per seed).
Our PSG-JEPA outperforms all compared baselines.
}
\label{tab:policy}
\end{wraptable}

We evaluate policy learning on LIBERO-Goal (10 goal-directed manipulation
tasks)~\cite{liu2023libero}. For each method, we train a task-conditioned OFT~\cite{li2026predictive} action head while jointly
fine-tuning the encoder. The head has hidden width 1024 and four layers, takes two visual frames as
input, and predicts action chunks of eight steps. We train for 30 epochs and report mean \(\pm\) std over
3 seeds, with 50 rollouts per task and seed (500 per seed, \(1{,}500\) in total).

\paragraph{Policy performance on LIBERO-Goal.}
As shown in Table~\ref{tab:policy}, our PSG-JEPA reaches \(85.3\%\) mean success
versus \(77.7\%\) for LeWM (\(+7.6\)), and also exceeds \aidm{}
(\(82.6\%\)) and DINOv2 (\(80.1\%\)) under the same policy-learning protocol.
This result extends the planning results in Sec.~\ref{sec:planning}: physically
grounded representations benefit not only goal-conditioned planners but also
task-conditioned policy learning across the ten LIBERO-Goal tasks. The
comparisons with \aidm{} and DINOv2 further show that neither action-level
supervision nor large-scale visual pretraining alone provides the same benefit
for this policy-learning setting.

\subsubsection{Real-World Robot.}
We further evaluate PSG-JEPA on a physical dual-arm Cobot (AgileX
Robotics) that adopts the Mobile ALOHA system design~\cite{fu2024mobile}. Each arm has six revolute joints
and a parallel gripper, and policies act from RGB observations captured
by three cameras: a front view and two wrist views (see Fig.~\ref{fig:real_robot}(a)).
We study three tabletop manipulation tasks (\emph{Place-to-Bread},
\emph{Place-to-Plate}, and \emph{Pour-Water}) and collect \(100\) teleoperated
demonstrations per task. Following the same policy-learning protocol as in
simulation, we adapt each representation with the shared action head and report
the average success rate over \(50\) trials per task.

As shown in Fig.~\ref{fig:real_robot}(b), our PSG-JEPA outperforms the LeWM baseline
on all three tasks: Place-to-Bread (\(84\%\) vs.\ \(62\%\)), Place-to-Plate
(\(74\%\) vs.\ \(58\%\)), and Pour-Water (\(80\%\) vs.\ \(60\%\)), i.e.,
\(79.3\%\) versus \(60.0\%\) on average. These results confirm that physically grounded
representations extend effectively to real-world manipulation.

\subsection{Component Ablations (Q4)}
\label{sec:ablation}

To validate the effectiveness of the key components of our PSG-JEPA, we conduct an
ablation study on OGBench-Cube and LIBERO-Goal. We evaluate each variant across
latent identifiability, goal-conditioned planning, and policy
learning. Results are summarized in Table~\ref{tab:ablation_unified}.
The \textbf{w/o transition grounding} variant retains only the single-latent state head, and
\textbf{w/o state grounding} removes the state head while retaining transition grounding.
The \textbf{w/o multi-horizon (adjacent)} and \textbf{w/o multi-horizon (endpoint)} variants replace the
multi-horizon joint-angle changes with only the adjacent (\(k{=}1\)) or the endpoint
(first\(\to\)last) horizon, respectively.

\begin{wraptable}{r}{0.48\textwidth}
\centering
\captionsetup{font=footnotesize}
{\scriptsize
\setlength{\tabcolsep}{0.55mm}
\resizebox{\linewidth}{!}{%
\begin{tabular}{lcc c c}
\toprule
 & \multicolumn{2}{c}{Probe (mean lin.\ \(r\))} & Planning & Policy \\
\cmidrule(lr){2-3}\cmidrule(lr){4-4}\cmidrule(lr){5-5}
Variant & state & transition & SR@5ep & LIBERO \\
\midrule
LeWM & 0.68 & 0.62
  & 80.7\(\pm\)1.9 & 77.7\(\pm\)0.5 \\
w/o transition    & 0.93 & 0.72 & 81.3\(\pm\)2.4 & 80.3\(\pm\)3.9 \\
w/o state         & 0.69 & 0.67 & 93.3\(\pm\)1.2 & 80.0\(\pm\)2.3 \\
w/o multi-horizon (adj.) & 0.86 & 0.74 & 93.5\(\pm\)1.9 & 81.5\(\pm\)3.1 \\
w/o multi-horizon (end.) & 0.86 & \textbf{0.75} & 93.7\(\pm\)2.0 & 81.2\(\pm\)2.7 \\
\midrule
\textbf{PSG-JEPA (full)} & \textbf{0.94} & \textbf{0.75} & \textbf{95.0\(\pm\)0.7} & \textbf{85.3\(\pm\)3.9} \\
\bottomrule
\end{tabular}%
}
}
\caption{
  \textbf{Component ablations across identifiability, planning, and policy learning.}
  State and transition report mean linear-probe \(r\), i.e., robot proprioception
  and transition identifiability. Planning reports success rate (\%) on OGBench-Cube
  (mean \(\pm\) std over 3 planner seeds), and policy reports success rate (\%) on
  LIBERO-Goal (mean \(\pm\) std over 3 seeds). Full PSG-JEPA
  attains the best overall performance across the three evaluation levels.
  }
\label{tab:ablation_unified}
\end{wraptable}

The ablations reveal distinct sensitivity patterns across the three evaluations.
Removing transition grounding produces the largest planning drop
(\(95.0\to81.3\)), whereas removing state grounding causes the largest drop in
mean linear state-probe performance (\(0.94\to0.69\)). On LIBERO-Goal, every
ablated variant lowers policy success to roughly \(80\)--\(81.5\%\), compared
with \(85.3\%\) for the full model. The differences among the ablated
variants are within seed noise. On single-task Cube planning, full multi-horizon grounding gives a modest
improvement over the adjacent- and endpoint-only variants (\(95.0\pm0.7\) versus
\(93.5\pm1.9\) and \(93.7\pm2.0\)), and a larger mean gap on policy
learning (\(85.3\) versus \(81.5\)). Overall, every grounding component
contributes, and the full model is best across all three evaluation levels.

\section{Conclusion}
In this work, we introduced PSG-JEPA, an end-to-end JEPA world model that grounds individual
latents in robot proprioceptive state and latent pairs in multi-horizon joint-angle changes. The
grounding heads are used only during training, so deployment retains the original JEPA inference
cost, while the grounding objectives make state and state-change information more
readily identifiable from individual latents and latent pairs, respectively.
Across latent identifiability, goal-conditioned planning, and policy
learning, our PSG-JEPA achieves the best overall performance among all compared methods.
Ablations show distinct sensitivity patterns: removing state grounding most
strongly reduces state identifiability, removing transition grounding
produces the largest planning drop, and the full objective achieves the
highest multi-task policy success. Together, these results show that
physical grounding improves both the identifiability of latent representations
and their usefulness for embodied decision-making.

\bibliographystyle{assets/plainnat}
\bibliography{references}

\end{document}